\documentclass{article}

\RequirePackage{natbib}
\usepackage{graphicx} 
\usepackage[utf8]{inputenc} 
\usepackage[T1]{fontenc}    
\usepackage{hyperref}       
\usepackage{url}            
\usepackage{booktabs}       
\usepackage{amsfonts}       
\usepackage{nicefrac}       
\usepackage{microtype}      
\usepackage{listings}       
\usepackage{mathptmx}       

\usepackage[margin=0.5cm]{caption} 

\newcommand{\specialcellcenter}[2][c]{%
  \begin{tabular}[#1]{@{}c@{}}#2\end{tabular}}

\setcitestyle{authoryear,round,citesep={;},aysep={,},yysep={;}}

\renewcommand{\cite}[1]{\citep{#1}}

\usepackage{blindtext}
\usepackage{geometry}
\begin{document}

\lstset{basicstyle=\footnotesize\ttfamily, language=Python, columns=flexible, breaklines=true, breakatwhitespace=true, stepnumber=0, numbersep=8pt, tabsize=4, showspaces=False, showstringspaces=False,}

{\fontfamily{ptm}\selectfont
\title{Gini Coefficient as a Unified Metric for Evaluating \\ Many-versus-Many Similarity in Vector Spaces}}

\author{Ben Fauber\thanks{Correspondence to: Ben.Fauber@dell.com} \\
\normalsize{Dell Technologies}
}
\date{November 4, 2024}

\maketitle

\begin{abstract}
We demonstrate that Gini coefficients can be used as unified metrics to evaluate many-versus-many (all-to-all) similarity in vector spaces. Our analysis of various image datasets shows that images with the highest Gini coefficients tend to be the most similar to one another, while images with the lowest Gini coefficients are the least similar. We also show that this relationship holds true for vectorized text embeddings from various corpuses, highlighting the consistency of our method and its broad applicability across different types of data. Additionally, we demonstrate that selecting machine learning training samples that closely match the distribution of the testing dataset is far more important than ensuring data diversity. Selection of exemplary and iconic training samples with higher Gini coefficients leads to significantly better model performance compared to simply having a diverse training set with lower Gini coefficients. Thus, Gini coefficients can serve as effective criteria for selecting machine learning training samples, with our selection method outperforming random sampling methods in very sparse information settings.
\end{abstract}

\section{Introduction}

Similarity in vector spaces quantifies how closely two or more vectors resemble each other. Similarity is often quantified using metrics such as cosine similarity, dot product, or Euclidean distance \cite{strang2014linear}. These measures are crucial for machine learning objectives like clustering, classification, and recommendation systems, as they help describe relationships between data points within their vector space \cite{strang2019linear, hardtrecht2022patterns}.

Similarity search is an important one-versus-many method used to find items in a dataset that are similar to a query item \cite{10.5555/1951721}. Yet, there is not a sufficient metric to assess the quality of many-versus-many similarity. For example, there is an insufficient method and metric to determine how similar all vectors in a dataset are with one another. Herein, we propose the use of the Gini coefficient as a unified metric for assessing many-versus-many similarity in vector space (Figure 1).

\begin{figure}[ht]
\begin{center}
\includegraphics[width=40mm]{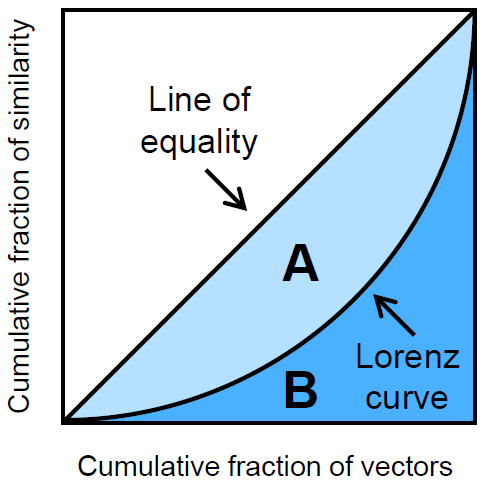}
\caption{Illustration of our proposal: the Gini coefficient can serve as a unified, singular metric to assess the many-versus-many similarity of vectors. The calculation involves the ratio between the area under the line of equality $A$ (light blue) and the area under the Lorenz curve $B$ (dark blue). The Gini coefficient for many vectors is determined by the ratio of these areas: $A/(A+B)$.}
\label{overview}
\end{center}
\end{figure}

\section{Background}

Similarity search is an fundamental one-versus-many search method used to find items in a dataset that are similar to a query item \cite{10.5555/1951721}. It is widely used in fields like image and video search, natural language processing, and healthcare to find similar instances. 

Similarity search involves representing/embedding $n$ items as $n$ vectors $\mathbb{R}^d$ in a high-dimensional space $d$ that accurately captures their semantic meaning. An exact match similarity search is of $O(n)$ complexity, where $n$ is the size of the dataset being queried. Approximate nearest neighbor (ANN) search algorithms such as locality-sensitive hashing (LSH) \cite{10.1145/276698.276876}, Min-hash \cite{10.1145/276698.276781}, product quantization (PQ) \cite{PQ5432202}, and hierarchical navigable small-world (HNSW) graphs \cite{malkov2018efficientrobustann} enable efficient queries of large datasets with sublinear time complexity \cite{andoni2018ann}. Despite the impressive advances in approximate nearest neighbor search algorithms, there is a lack of suitable methods and metrics to describe the similarity of all vectors in a dataset with one another.

The Gini coefficient, also known as the Gini index or Gini ratio, is typically applied in economic studies to measure income, wealth, or consumption inequality within a country or group. Originally proposed by the Italian statistician Corrado Gini, and built on the work of American economist Max Lorenz, Gini's method proposed the difference between a hypothetical line of perfect equality and the actual income distribution line as a measure of inequality \cite{gini1912variabilita, RePEc:mtn:ancoec:0501, RePEc:tur:wpapnw:070}.

In addition to its original application in economics, the Gini coefficient can be applied in any field of science that studies distributions. For example, in ecology the Gini coefficient has been used as a measure of biodiversity \cite{WittebolleNature2009}. In public health, it has been used as a measure of health-related quality of life within a population \cite{AsadaHealthQuality2005}. In medicinal chemistry, it has been used to express the selectivity of small molecule protein kinase inhibitors against a broader panel of protein kinases \cite{GraczykGiniKinasese2007}.

When applied to economic equity, the Gini coefficient ranges from 0 (perfect equality, where everyone has the same income) to 1 (perfect inequality, where one person has all the income). To facilitate the calculation, the income data is arranged in ascending order. Next, the cumulative shares of the population and their corresponding incomes are calculated. Plotting the cumulative shares versus the cumulative fraction/count results in the Lorenz curve, which illustrates the income distribution (Figure 1). The Gini coefficient is determined by the area between the Lorenz curve and the line of equality.

\subsection{Our Contribution}

We propose that the Gini coefficient can be used as a unified, singular metric to assess the many-versus-many similarity of multiple vectors (Figure 1). In our setting, a set of $n$ real value vectors $\mathbf{v} \in \mathbb{R}^d$ of $d$ dimensions can be $\ell_2$-normalized and represented as a matrix $\mathbf{V} \in \mathbb{R}^{n \times d}$ of $n$  vectors. Further, the scalar product $\mathbf{S}$ of $\mathbf{V}\mathbf{V}^{\top}$ results in a similarity matrix $\mathbf{S} \in \mathbb{R}^{n \times n}$, where each $\mathbf{s}_{ij} \in \mathbf{S}$ represents the similarity between vectors $\mathbf{v}_i$ and $\mathbf{v}_j$ and $\mathbf{s}_{ij} \in [-1,1]$.

Calculating the Gini coefficient $\mathbf{g}_{i}$ associated with row $\mathbf{s}_i \in \mathbf{S}$ results in a single value metric that represents the similarity for $\mathbf{v}_i$ versus all other $\mathbf{v}_{n}$. Specifically, the Gini coefficient $\mathbf{g}_{i} \in \mathbf{G}$, where $\mathbf{G} \in \mathbb{R}^n$, for $\mathbf{s}_{i} \in \mathbf{S}$ is calculated as the area under the curve (AUC) below the line of equality (Figure 1, area $A$), divided by the sum of the AUC below the line of equality and the AUC below the Lorenz curve (Figure 1, area $A+B$) \cite{GraczykGiniKinasese2007}. Calculation of the Gini coefficients of $\mathbf{s}_{ij} \in \mathbf{S}$ results in $\mathbf{G} \in \mathbb{R}^n$. The resulting Gini coefficients $\mathbf{G}$ can be normalized with a $MinMax$ scaling approach such that $\mathbf{g}_{i} \in [0,1], \forall \mathbf{g}_{i} \in \mathbf{G}$.\footnote{https://scikit-learn.org/1.5/modules/generated/sklearn.preprocessing.MinMaxScaler.html (accessed 14Oct2024).}

In our application, the Gini coefficient $\mathbf{g}_{i}$ was bounded $[0,1]$, but the relevance of the value to the outcome will vary based on the objective. In our assessment of many-versus-many (\emph{e.g.}, all-to-all) similarity, higher Gini coefficients represented greater similarity, whereas lower Gini coefficients represented lesser similarity.

We also demonstrated that Gini coefficients can guide data sampling strategies when training machine learning (ML) algorithms in very sparse information settings. In multiclass classification, when the training and testing dataset distributions were closely aligned, prioritizing the top one or two samples with the highest Gini coefficients resulted in the selection of exemplary and prototypical samples from each class. This approach outperformed random selection of training samples and was particularly valuable when only one or two samples per class were used for multiclass ML training campaigns.

\section{Results}

The MNIST (Modified National Institute of Standards and Technology database) dataset is a collection of 70,000 handwritten digit images, each 28x28 pixels in size, grayscale, and generally used for training and testing machine learning models.\footnote{https://pytorch.org/vision/stable/generated/torchvision.datasets.MNIST.html (accessed 14Oct2024).} It includes 60,000 training images and 10,000 test images, with an approximately uniform distribution of all digits 0 through 9, and is a standard benchmark for image recognition tasks.

\begin{figure}[ht]
\begin{center}
\includegraphics[width=60mm]{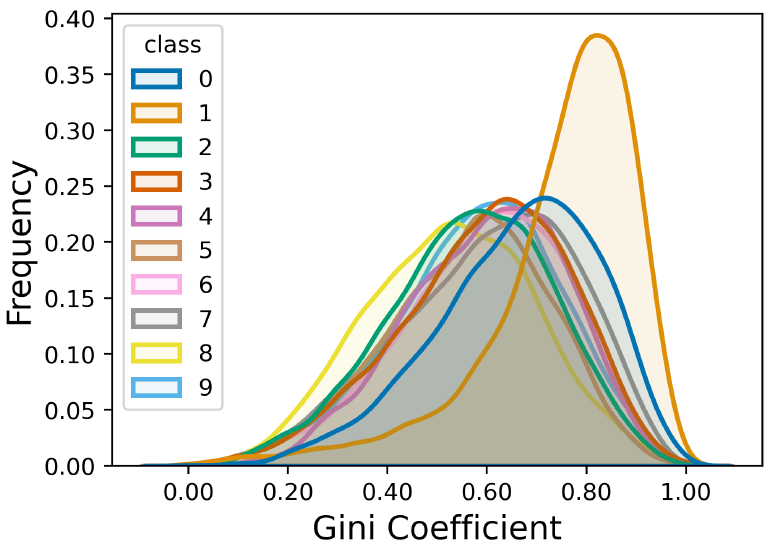}
\caption{Kernel density plot showing the Gini coefficients for each class in the MNIST training dataset, which contains 60,000 instances in total and approximately 6,000 instances per class. The Gini coefficients were calculated using the flattened raw pixel values ($d = 784$) ranging from 0 to 255 for each $28 \times 28$ grayscale image. Each $d$-dimensional image vector was $\ell_2$-normalized before computing the similarity values and Gini coefficients. The Gini coefficients were $MinMax$ normalized $[0,1]$ to allow for comparison across classes.}
\label{gini-by-class}
\end{center}
\vskip -0.2in
\end{figure}

\subsection{Gini Coefficients as Metrics for Many-versus-Many Similarity}

\subsubsection{MNIST Dataset}

Our method was applied to the MNIST dataset to explore the impact of the Gini coefficient as a unified metric for evaluating many-versus-many similarity in vector spaces. Specifically, the per-class cosine similarities of the flattened ($d = 784$) and $\ell_2$-normalized raw pixel grayscale MNIST dataset images were calculated. All elements $\mathbf{s}_{ij} \in \mathbf{S}$ of the resulting similarity matrix $\mathbf{S}$ were subtracted from 1 such that the most similar elements in $\mathbf{S}$ were $\mathbf{s}_{ij} \sim 0$. The per-class Gini coefficients for $\mathbf{S}$ were then calculated and $MinMax$ normalized\textsuperscript{1} to ensure all values were $[0,1]$ and to allow for comparison of Gini coefficients across all classes in the dataset.

\begin{figure}[ht]
\begin{center}
\includegraphics[width=100mm]{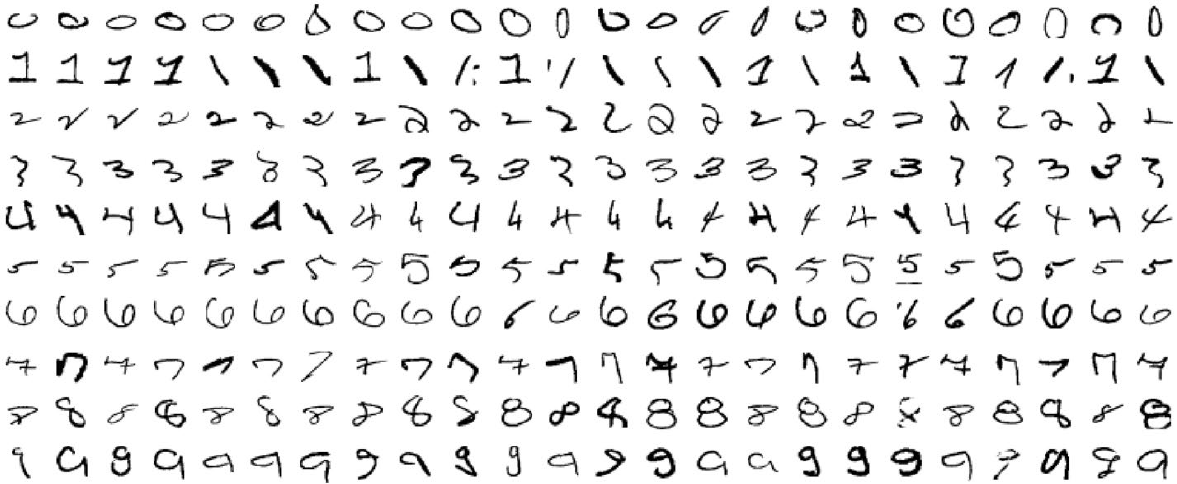}
\caption{Top-24 examples with the \emph{lowest} Gini coefficients for each class in the MNIST training dataset, which contains 60,000 instances in total and approximately 6,000 instances per class. The Gini coefficients were calculated using the flattened raw pixel values ($d = 784$) ranging from 0 to 255 for each $28 \times 28$ grayscale image. Each $d$-dimensional image vector was $\ell_2$-normalized before computing the similarity values and Gini coefficients. The Gini coefficients were $MinMax$ normalized $[0,1]$ to allow for comparison across classes.}
\label{mnist-lowest-gini}
\end{center}
\end{figure}

The kernel density plot in Figure 2 illustrates the $MinMax$ normalized Gini coefficients for all 10 classes in the MNIST dataset. Most classes showed an approximate normal distribution of their Gini coefficients, whereas the classes "1" and "0" showed Poisson distributions. Class "1" exhibited the strongest Poisson distribution of its Gini coefficients, highlighting the similarity of "1" images across the dataset. This was likely because the number one is usually drawn with one or a few strokes of a pencil or pen, leading to limited variation. Similarly, the number zero exhibited high similarity values, as indicated by its higher Gini coefficients, because there are only so many ways to draw the number. In contrast, other numbers exhibited greater variation in Gini coefficients due to the increased complexity of their depictions compared to zero and one.

\begin{figure}[ht]
\begin{center}
\includegraphics[width=100mm]{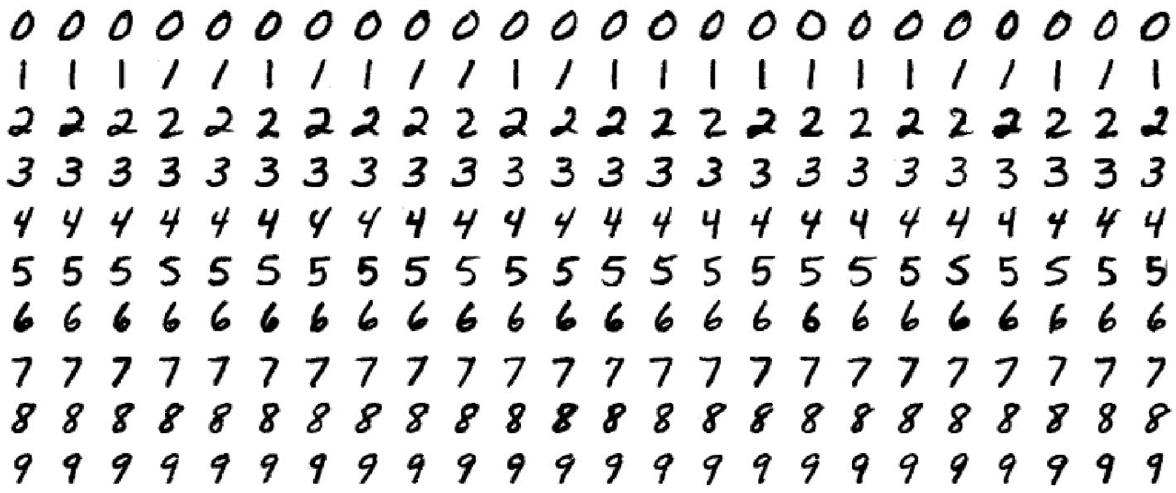}
\caption{Top-24 examples with the \emph{highest} Gini coefficients for each class in the MNIST training dataset, which contains 60,000 instances in total and approximately 6,000 instances per class. The Gini coefficients were calculated using the flattened raw pixel values ($d = 784$) ranging from 0 to 255 for each $28 \times 28$ grayscale image. Each $d$-dimensional image vector was $\ell_2$-normalized before computing the similarity values and Gini coefficients. The Gini coefficients were $MinMax$ normalized $[0,1]$ to allow for comparison across classes.}
\label{mnist-highest-gini}
\end{center}
\end{figure}

The utility of the Gini coefficient is demonstrated in Figure 3, which displays the top 24 images of each class in the MNIST dataset with the \emph{lowest} per-class Gini coefficients. In this context, lower Gini coefficients indicated less similarity compared to higher coefficients. Therefore, the examples in Figure 3 represented the most unique images for each class in the dataset.

Conversely, the highest per-class Gini coefficients in the MNIST dataset highlight the most similar images (Figure 4). These highly similar images, with the highest per-class Gini coefficients, can be considered exemplary of each class in the dataset. Therefore, the images in Figure 4 represented the most common depictions of each number in the dataset.

\begin{figure}[hb]
\begin{center}
\includegraphics[width=100mm]{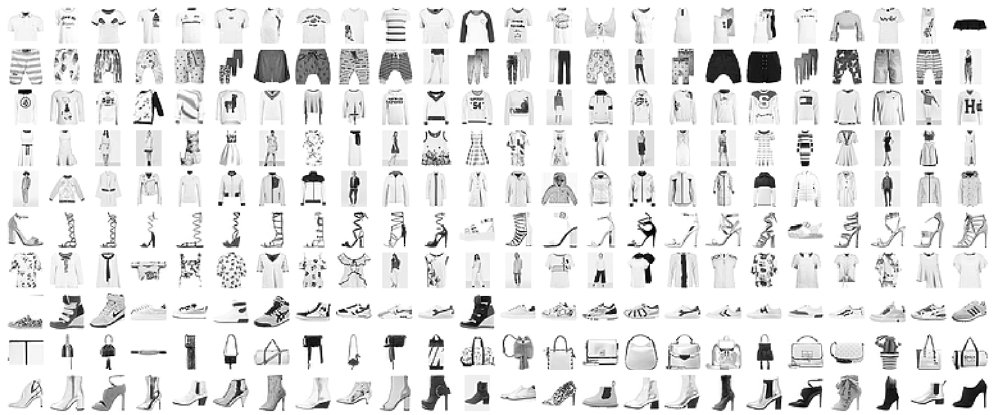}
\caption{Top-24 examples with the \emph{lowest} Gini coefficients for each class in the Fashion-MNIST training dataset, which contains 60,000 instances in total and approximately 6,000 instances per class. The Gini coefficients were calculated using the flattened raw pixel values ($d = 784$) ranging from 0 to 255 for each $28 \times 28$ grayscale image. Each $d$-dimensional image vector was $\ell_2$-normalized before computing the similarity values and Gini coefficients. The Gini coefficients were $MinMax$ normalized $[0,1]$ to allow for comparison across classes.}
\label{fashionmnist-lowest-gini}
\end{center}
\end{figure}

\subsubsection{Fashion-MNIST Dataset}

Additionally, our method was applied to the Fashion-MNIST dataset. Fashion-MNIST is a dataset containing 70,000 unique grayscale images of clothing, where each image is $28 \times 28$ pixels in size. The dataset includes 10 classes of clothing items, which are approximately evenly distributed: T-shirts, trousers, pullovers, dresses, coats, sandals, shirts, sneakers, bags, and ankle boots.\footnote{https://pytorch.org/vision/0.19/generated/torchvision.datasets.FashionMNIST.html (accessed 14Oct2024).} Similar to MNIST, Fashion-MNIST contains a predefined training set of 60,000 examples and a test set of 10,000 examples. The same approach and data preparation steps that we used with the MNIST dataset were applied to the Fashion-MNIST dataset to generate the $MinMax$ normalized per-class Gini coefficients $[0,1]$ for each class in the Fashion-MNIST training dataset.

\begin{figure}[ht]
\begin{center}
\includegraphics[width=100mm]{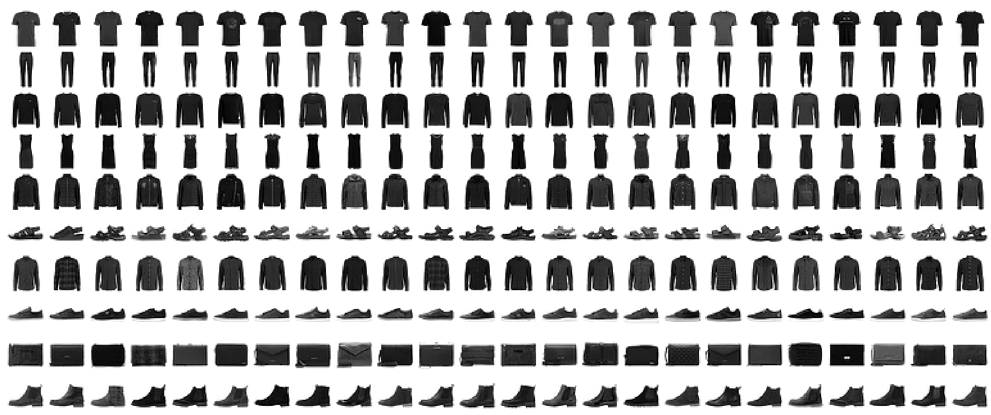}
\caption{Top-24 examples with the \emph{highest} Gini coefficients for each class in the Fashion-MNIST training dataset, which contains 60,000 instances in total and approximately 6,000 instances per class. The Gini coefficients were calculated using the flattened raw pixel values ($d = 784$) ranging from 0 to 255 for each $28 \times 28$ grayscale image. Each $d$-dimensional image vector was $\ell_2$-normalized before computing the similarity values and Gini coefficients. The Gini coefficients were $MinMax$ normalized $[0,1]$ to allow for comparison across classes.}
\label{fashionmnist-highest-gini}
\end{center}
\end{figure}

The results of the Gini coefficient analysis with the Fashion-MNIST dataset were like those of the MNIST dataset. Namely, the images with the lowest per-class Gini coefficients were the most unique images in each class (Figure 5). Conversely, the images with the highest per-class Gini coefficients were the most similar and most exemplary images in each class (Figure 6). These observations aligned with our findings from the MNIST dataset, and it was encouraging to see the usefulness of the Gini coefficient demonstrated across these different image datasets.

\begin{figure}[hb]
\begin{center}
\includegraphics[width=100mm]{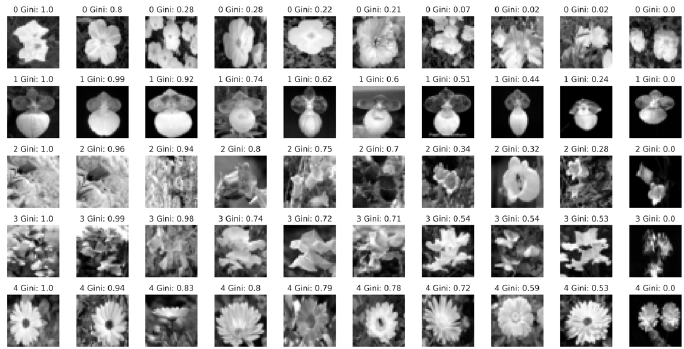}
\caption{Gini coefficients by class, from highest to lowest (left to right), for five classes (classes 0-4) in the Flowers102 training dataset. Each class in the dataset contained 10 images with a total of 102 distinct classes. The Gini coefficients were calculated using the flattened raw pixel values ($d = 1,296$) ranging from 0 to 255 for each $36 \times 36$ grayscale image. Each $d$-dimensional image vector was $\ell_2$-normalized before computing the similarity values and Gini coefficients. The Gini coefficients were $MinMax$ normalized $[0,1]$ to allow for comparison across classes.}
\label{flowers-gini-5classes}
\end{center}
\end{figure}

\begin{table*}[ht]
\begin{center}
\begin{small}
\begin{tabular}{cclc}
\toprule
\specialcellcenter{Text Source} &
\specialcellcenter{Author} &
\specialcellcenter{Text Chunk} &
\specialcellcenter{Gini Coefficient} \\
\midrule
\emph{The Trial} & F. Kafka & "fleas in the doorkeeper's fur collar" & 0 \\
\emph{The Trial} & F. Kafka & "grey constructions, tall blocks of flats" & 0.03 \\
\emph{The Trial} & F. Kafka & "picture of the Virgin Mary, an" & 0.05 \\
\emph{The Trial} & F. Kafka & "plate hung on the wall, casting" & 0.05 \\
\emph{The Trial} & F. Kafka & "a guide to the city's tourist" & 0.06 \\
\midrule
\emph{The Trial} & F. Kafka & "hesitated he went over to the" & 0.94 \\
\emph{The Trial}  & F. Kafka & "it to him. As he looked" & 0.94 \\
\emph{The Trial} & F. Kafka & "it over as if he had" & 0.95 \\
\emph{The Trial} & F. Kafka & "to his left--and this was the" & 0.99 \\
\emph{The Trial} & F. Kafka & "him as he stepped through their" & 1 \\
\midrule
\emph{A Study in Scarlet} & A. C. Doyle & "Literature.—Nil. 2. Philosophy.—Nil. 3. Astron..." & 0 \\
\emph{A Study in Scarlet} & A. C. Doyle & "Van Jansen, in Utrecht, in the" & 0 \\
\emph{A Study in Scarlet} & A. C. Doyle & "Joseph Stangerson? The Secretary, Mr. Joseph" & 0.03 \\
\emph{A Study in Scarlet} & A. C. Doyle & "hotels and lodging-houses in the vicinity" & 0.03 \\
\emph{A Study in Scarlet} & A. C. Doyle & "pounds thirteen. Pocket edition of Boccaccio’s" & 0.04 \\
\midrule
\emph{A Study in Scarlet} & A. C. Doyle & "which he had set himself to" & 0.96 \\
\emph{A Study in Scarlet} & A. C. Doyle & "upon his, and looking up, he" & 0.98 \\
\emph{A Study in Scarlet} & A. C. Doyle & "out to his wheatfields, when he" & 0.99 \\
\emph{A Study in Scarlet} & A. C. Doyle & "beside him, holding on to the" & 0.99 \\
\emph{A Study in Scarlet} & A. C. Doyle & "to what he takes into his" & 1 \\
\bottomrule
\end{tabular}
\end{small}
\caption{Top-5 examples of the lowest and highest Gini coefficients associated with six-word text chunks from a parent text source. A vectorized embedding ($d = 512$) was created for each text chunk with the Universal Sentence Encoder v4 (USE). All text embeddings were $\ell_2$-normalized prior to similarity assessment and Gini coefficients for each text source were $MinMax$ normalized to [0, 1].}
\end{center}
\end{table*}

\subsubsection{Flowers102 Dataset}

Application of our method to the Flowers102 dataset resulted in clear ranking of most-similar to least-similar images for each class in the dataset. Flowers102 is a dataset containing 1,020 unique training images of flowers from 102 different classes, with 10 images per class.\footnote{https://pytorch.org/vision/0.19/generated/torchvision.datasets.Flowers102.html (accessed 14Oct2024).} The same approach and data preparation steps that we used with the MNIST dataset were applied to the Flowers102 dataset, in addition to converting the training images to grayscale and resizing to $36 \times 36$ pixels, to generate the $MinMax$ normalized per-class Gini coefficients $[0,1]$ for each class in the Flowers102 dataset. The results of the Flowers102 Gini coefficient analysis clearly demonstrated that the most similar images in each class had highest Gini coefficients whereas the lowest Gini coefficients were the most unique images for each class (Figure 7).

\subsubsection{Text Analysis}

Our method was also applied to vectorized text embeddings to assess the utility of Gini coefficients when comparing text chunks. In our analysis, two novels\footnote{The text was sourced from Project Gutenberg: https://www.gutenberg.org/ (accessed 14Oct2024).} were each chunked into six-word chunks and the six-word chunks from each novel were separately embedded into $512$-dimensional vectors using the Universal Sentence Encoder v4 (USE) \cite{googlegusecer}. The resulting $\ell_2$-normalized vectors were used in our method, all Gini coefficients for each novel were $MinMax$ normalized to [0,1], and the outcomes for each novel are shown in Table 1.

Notably, the text chunks in Table 1 with the lowest Gini coefficients were the most unique text phrases from each novel, containing descriptive and colorful nouns and adjectives. Conversely, the text chunks with the highest Gini coefficients were the least unique and bland text phrases containing common pronouns, possessives, conjunctions, and interrogative words. At a high level, the Gini coefficient patterns observed with our text data analysis matched the Gini coefficient patterns we identified with the MNIST, Fashion-MNIST, and Flowers102 image datasets. These results further emphasized the reliability of our method and the wide applicability of the Gini coefficient for evaluating many-to-many similarity, regardless of the data types involved.

\subsection{Gini Coefficients can Guide Dataset Sampling for Machine Learning}

Statistical sampling and randomization are crucial in statistics, especially for designing experiments and surveys \cite{ISL2013}. Randomization, which involves randomly assigning elements to different groups or treatments, is vital for minimizing bias, controlling for variables, and ensuring validity. 

Most canonical machine learning (ML) pipelines and processes utilize randomized selection of data for training and testing to improve robustness and minimize bias \cite{AGeronMLBook, raschka2022machine}. Recently, the notion of learning in sparse information settings has gained attention, particularly the concept of data pruning \cite{okanovic2023repeatedrandomsamplingminimizing, paul2023deeplearningdatadiet, sorscher2023neuralscalinglawsbeating}. These efforts critically evaluate the improvements in deep learning algorithms that have been driven in part by increasingly larger datasets. Specifically, they aim to identify important examples in datasets for learning generalization while eliminating superfluous data.

We show that Gini coefficients can help guide data sampling strategies when training ML algorithms in very sparse information settings. To demonstrate this, we assessed the influence of Gini coefficients on the prioritization of MNIST training dataset samples for training a support vector machine (SVM) multiclass classification algorithm. We then evaluated the impact of the sampling strategy on the classification accuracy of the 10,000-example MNIST test dataset.

In our study, the training dataset samples were selected from the 60,000-example MNIST training dataset by either random sampling, prioritizing the samples with the highest per-class Gini coefficients (\emph{i.e.}, most exemplary and least diverse samples), or prioritizing the samples with the lowest per-class Gini coefficients (\emph{i.e.}, most diverse samples). The mean of three random sampling results were used to generate each random sampling data point. All MNIST training and testing data inputs to the SVM model were flattened raw pixel values ($d = 784$) ranging from 0 to 255 for each $28 \times 28$ grayscale image. Each $d$-dimensional image vector was $\ell_2$-normalized before entering the SVM model for either training or testing (\emph{see} Appendix for details).

We noted that when selecting only one or two exemplary samples per class, prioritizing the samples with the highest per-class Gini coefficients outperformed all other sampling strategies (Figure 8). Conversely, prioritizing the samples with the lowest per-class Gini coefficients resulted in significantly lower performance than the other sampling methods. In aggregate, the random sampling approach outperformed the other methods and led to near-optimal performance of the SVM classifier with undersampled training dataset examples. All methods converged to the maximum accuracy of 92\% as they neared the maximum number of approximately 6,000 samples per class of the MNIST training data set due to the Central Limit Theorem \cite{fullersamplingstatistics}.

\begin{figure}[ht]
\begin{center}
\includegraphics[width=100mm]{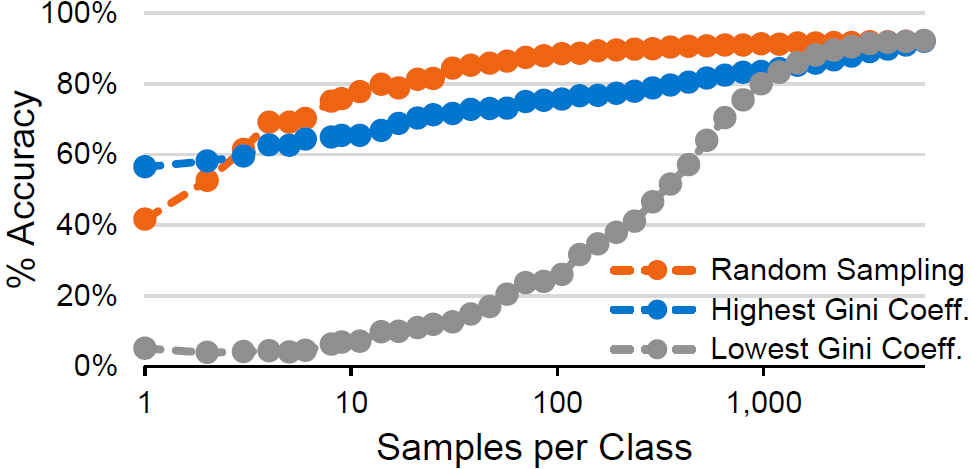}
\caption{Log-linear plot showing the accuracy of support vector machine (SVM) multiclass classification of the 10,000-example MNIST test dataset versus the number of samples per class from the 60,000-example MNIST training dataset used for training the SVM. The training dataset samples were selected by either per-class random sampling (orange), prioritizing the samples with the highest per-class Gini coefficients (blue), or prioritizing the samples with the lowest per-class Gini coefficients (gray). The mean of three random sampling results is shown for each "random sampling" data point (orange). All MNIST inputs to the SVM model were flattened raw pixel values (dimension $d=784$) ranging from 0 to 255 for each $28 \times 28$ grayscale image. Each $d$-dimensional image vector was $\ell_2$-normalized before entering the SVM model.}
\label{mnist-svm-accuracy}
\end{center}
\end{figure}

In multiclass classification, when the training and testing dataset distributions were closely aligned, prioritizing the top one or two samples with the highest Gini coefficients resulted in the selection of exemplary samples from each class. This approach outperformed random selection of training samples and was particularly valuable when only one or two samples per class are used for multiclass ML training campaigns.

At a high level, we observed that selecting machine learning training samples that closely matched the distribution of the testing dataset through random sampling (\emph{see} Appendix for details) was much more important than ensuring data diversity (\emph{i.e.}, samples with lower Gini coefficients). Selection of exemplary and iconic training samples (\emph{i.e.}, samples with higher Gini coefficients) led to significantly better model performance compared to simply having a diverse training set (\emph{i.e.}, samples with lower Gini coefficients).

Similarly, we applied this sampling analysis to the Fashion-MNIST dataset and found similar patterns and trends as observed with the MNIST dataset (\emph{see} Appendix for details). These consistent results, obtained using a different image dataset, further highlighted the effectiveness of our method in prioritizing samples with the highest per-class Gini coefficients, especially in situations with very sparse information.

\section{Conclusion}

We demonstrated that Gini coefficients can be used as a unified metric to evaluate many-versus-many (\emph{e.g.}, all-to-all) similarly in vector spaces. Our analysis of various image datasets showed that images with the highest Gini coefficients tend to be the most similar to one another, while those with the lowest Gini coefficients were the least similar. We also showed that this relationship holds true for vectorized text embeddings from various novels, highlighting the consistency of our method and its broad applicability across different types of data. 

Additionally, we demonstrated that selecting machine learning training samples that closely match the distribution of the testing dataset was far more important than ensuring data diversity. Selection of exemplary and iconic training samples led to significantly better model performance compared to simply having a diverse training set. Thus, Gini coefficients can serve as effective criteria for selecting machine learning training samples, with our selection method outperforming random sampling methods in very sparse information settings.

\newpage

\bibliography{bib-llms}
\bibliographystyle{icml2020}

\newpage

\appendix

\setcounter{table}{0}
\renewcommand{\thetable}{A\arabic{table}}

\setcounter{figure}{0}
\renewcommand{\thefigure}{A\arabic{figure}}

\section{Appendix}
\subsection{Gini Coefficient Calculation}

The results described in this article were carried out using open datasets on a Linux system configured with Anaconda v23.1.0. Additional python dependencies included: \texttt{tensorflow v2.16.2}, {tensorflow-hub v0.16.1}, 
\texttt{torch v2.1.1}, and \texttt{torchvision v0.20.1}. All processes can be run on CPU infrastructure, but the processes scale with minimal latency on GPU infrastructure. 

In order to calculate the Gini coefficients, a set of $n$ real value vectors $\mathbf{v} \in \mathbb{R}^d$ of $d$ dimensions were $\ell_2$-normalized and represented as a matrix $\mathbf{V} \in \mathbb{R}^{n \times d}$ of $n$  vectors. The scalar product $\mathbf{S}$ of $\mathbf{V}\mathbf{V}^{\top}$ resulted in a similarity matrix $\mathbf{S} \in \mathbb{R}^{n \times n}$, where each $\mathbf{s}_{ij} \in \mathbf{S}$ represented the similarity between vectors $\mathbf{v}_i$ and $\mathbf{v}_j$ and $\mathbf{s}_{ij} \in [-1,1]$. The similarity matrix $\mathbf{S}$ was subtracted from the scalar $1$ such that $\mathbf{s}_{ij} \approx 0$ represented the most similar $\mathbf{v}_i$ and $\mathbf{v}_j$.

Calculation of the Gini coefficients of $\mathbf{s}_{ij} \in \mathbf{S}$ resulted in $\mathbf{G} \in \mathbb{R}^n$. Calculation of the Gini coefficient $\mathbf{g}_{i}$ associated with row $\mathbf{s}_i \in \mathbf{S}$ was conducted in python according to the method of \cite{GraczykGiniKinasese2007}. The resulting Gini coefficients $\mathbf{G}$ were normalized with a $MinMax$ scaling approach such that $\mathbf{g}_{i} \in [0,1], \forall \mathbf{g}_{i} \in \mathbf{G}$. Thus, the calculation of $\mathbf{g}_{i}$ resulted in a single value metric that represented the similarity for $\mathbf{v}_i$ versus all other $\mathbf{v}_{n}$.  

In our application, the Gini coefficient $\mathbf{g}_{i}$ was bounded $[0,1]$, but the relevance of the value to the outcome will vary based on the objective. In our assessment of many-versus-many (\emph{e.g.}, all-to-all) similarity, higher Gini coefficients represented greater similarity, whereas lower Gini coefficients represented lesser similarity.

\newpage

\subsection{Gini Coefficients for MNIST Dataset Figures}

The MNIST training dataset contains 60,000 instances in total and approximately 6,000 instances per class. The Gini coefficients were calculated using the flattened raw pixel values ($d = 784$) ranging from 0 to 255 for each $28 \times 28$ grayscale image. Each $d$-dimensional image vector was $\ell_2$-normalized before computing the similarity values and Gini coefficients. The Gini coefficients were $MinMax$ normalized $[0,1]$ to allow for comparison across classes.

Figure A1 displays the top 24 images of each class in the MNIST dataset with the \emph{lowest} per-class Gini coefficients. Lower Gini coefficients indicated less similarity compared to higher coefficients. Therefore, the examples in Figure A1 represent the most unique images for each class in the dataset. Conversely, the highest per-class Gini coefficients in the MNIST dataset highlight the most similar images (Figure A2). These highly similar images, with the highest per-class Gini coefficients, can be considered exemplary of each class in the dataset.

\begin{figure}[ht]
\begin{center}
\includegraphics[width=140mm]{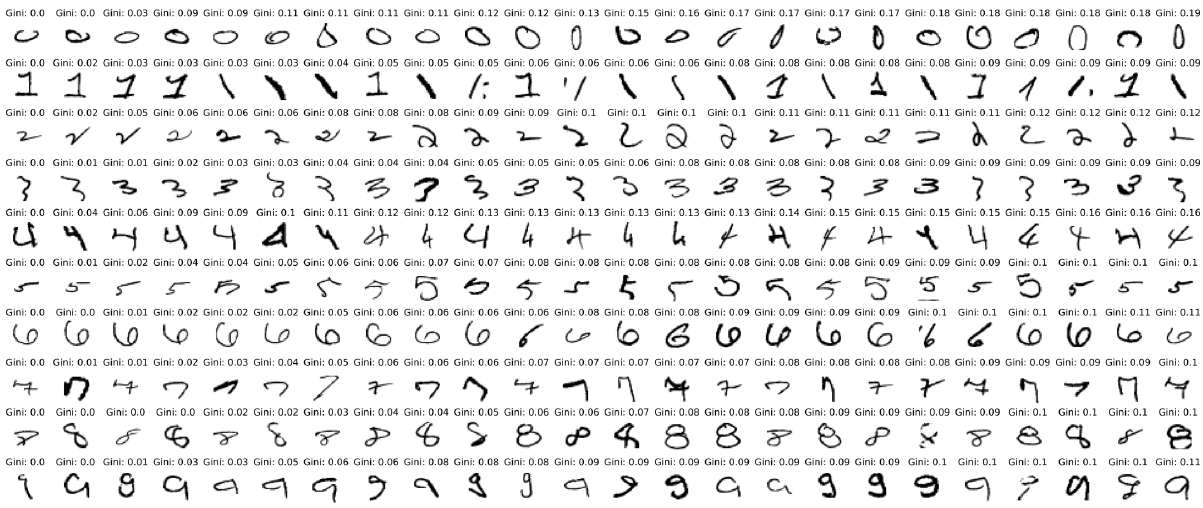}
\caption{Top-24 examples with the \emph{lowest} Gini coefficients for each class in the MNIST training dataset, which contains 60,000 instances in total and approximately 6,000 instances per class. The Gini coefficients are shown in the figure above each respective image. The values were calculated using the flattened raw pixel values ($d = 784$) ranging from 0 to 255 for each $28 \times 28$ grayscale image. Each $d$-dimensional image vector was $\ell_2$-normalized before computing the similarity values and Gini coefficients. The Gini coefficients were $MinMax$ normalized $[0,1]$ to allow for comparison across classes.}
\label{mnist-lowest-gini-with-values}
\end{center}
\vskip -0.2in
\end{figure}

\begin{figure}[ht]
\vskip -0.1in
\begin{center}
\includegraphics[width=140mm]{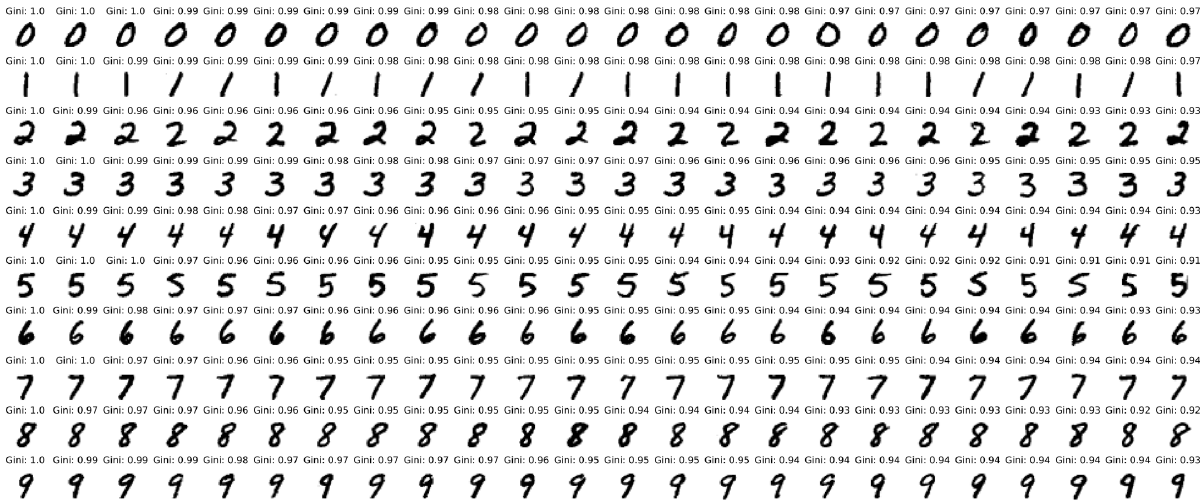}
\caption{Top-24 examples with the \emph{highest} Gini coefficients for each class in the MNIST training dataset, which contains 60,000 instances in total and approximately 6,000 instances per class. The Gini coefficients are shown in the figure above each respective image. The values were calculated using the flattened raw pixel values ($d = 784$) ranging from 0 to 255 for each $28 \times 28$ grayscale image. Each $d$-dimensional image vector was $\ell_2$-normalized before computing the similarity values and Gini coefficients. The Gini coefficients were $MinMax$ normalized $[0,1]$ to allow for comparison across classes.}
\label{mnist-highest-gini-with-values}
\end{center}
\vskip -0.3in
\end{figure}

\newpage

\subsection{Gini Coefficients for Fashion-MNIST Dataset Figures}

The Fashion-MNIST training dataset contains 60,000 instances in total and approximately 6,000 instances per class. The Gini coefficients were calculated using the flattened raw pixel values ($d = 784$) ranging from 0 to 255 for each $28 \times 28$ grayscale image. Each $d$-dimensional image vector was $\ell_2$-normalized before computing the similarity values and Gini coefficients. The Gini coefficients were $MinMax$ normalized $[0,1]$ to allow for comparison across classes.

Figure A3 displays the top 24 images of each class in the Fashion-MNIST dataset with the \emph{lowest} per-class Gini coefficients. Lower Gini coefficients indicated less similarity compared to higher coefficients. Therefore, the examples in Figure A3 represent the most unique images for each class in the dataset. Conversely, the highest per-class Gini coefficients in the Fashion-MNIST dataset highlight the most similar images (Figure A4). These highly similar images, with the highest per-class Gini coefficients, can be considered exemplary of each class in the dataset.

\begin{figure}[ht]
\begin{center}
\includegraphics[width=140mm]{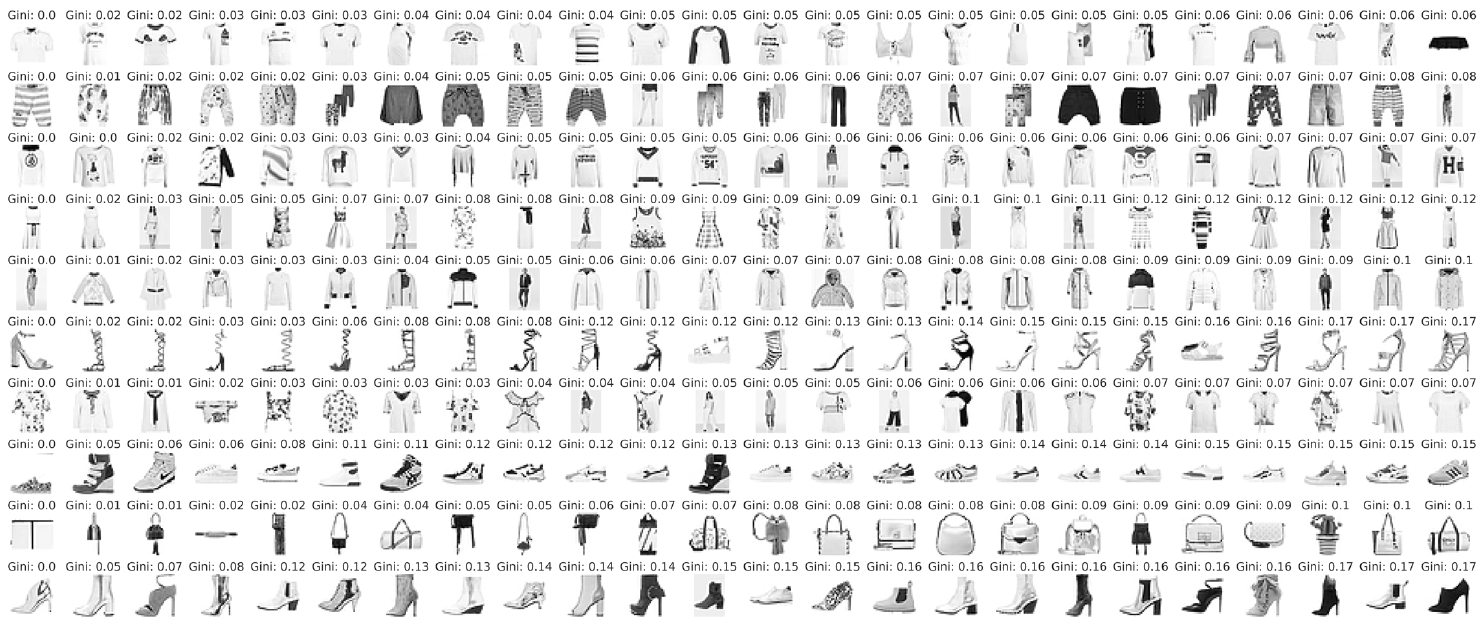}
\caption{Top-24 examples with the \emph{lowest} Gini coefficients for each class in the Fashion-MNIST training dataset, which contains 60,000 instances in total and approximately 6,000 instances per class. The Gini coefficients are shown in the figure above each respective image. The values were calculated using the flattened raw pixel values ($d = 784$) ranging from 0 to 255 for each $28 \times 28$ grayscale image. Each $d$-dimensional image vector was $\ell_2$-normalized before computing the similarity values and Gini coefficients. The Gini coefficients were $MinMax$ normalized $[0,1]$ to allow for comparison across classes.}
\label{fashionmnist-lowest-gini-with-values}
\end{center}
\vskip -0.2in
\end{figure}

\begin{figure}[ht]
\vskip -0.1in
\begin{center}
\includegraphics[width=140mm]{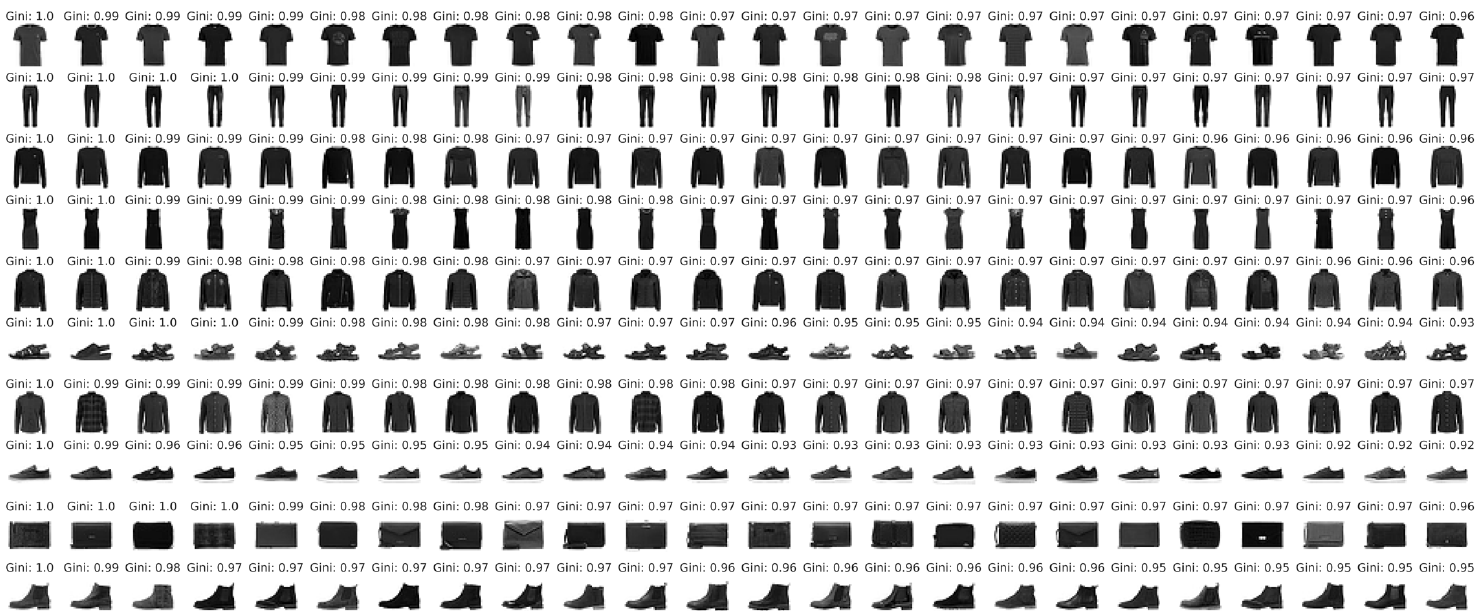}
\caption{Top-24 examples with the \emph{highest} Gini coefficients for each class in the Fashion-MNIST training dataset, which contains 60,000 instances in total and approximately 6,000 instances per class. The Gini coefficients are shown in the figure above each respective image. The values were calculated using the flattened raw pixel values ($d = 784$) ranging from 0 to 255 for each $28 \times 28$ grayscale image. Each $d$-dimensional image vector was $\ell_2$-normalized before computing the similarity values and Gini coefficients. The Gini coefficients were $MinMax$ normalized $[0,1]$ to allow for comparison across classes.}
\label{fashionmnist-highest-gini-with-values}
\end{center}
\vskip -0.3in
\end{figure}

\newpage

\subsection{Classifier Accuracies for MNIST Dataset Images}

We explored the role of sampling strategies when training a support vector machine (SVM) machine learning (ML) algorithm. Training datasets were sampled with increasing numbers of samples per class in the multiclass dataset. The resulting trained ML models were evaluated using a hold-out testing dataset that was consistent across the entire analysis. 

In our study, the training dataset samples were selected from the 60,000-example MNIST training dataset by either random sampling, prioritizing the samples with the highest per-class Gini coefficients (\emph{i.e.}, most exemplary and least diverse samples), or prioritizing the samples with the lowest per-class Gini coefficients (\emph{i.e.}, most diverse samples). The mean of three random sampling results were used to generate each random sampling data point. All MNIST training and testing data inputs to the SVM model were flattened raw pixel values ($d = 784$) ranging from 0 to 255 for each $28 \times 28$ grayscale image. Each $d$-dimensional image vector was $\ell_2$-normalized before entering the SVM model for either training or testing. SVM model performance was assessed by the classification accuracy across all classes for the test dataset for each training sample size cohort.

Follow-up analyses explored the impact of additional training sample selection methods. In this follow-up analysis, the training dataset samples were selected using one of the following methods: per-class random sampling, prioritizing samples with the highest per-class Gini coefficients, mimicking the per-class distribution of the test dataset Gini coefficients using kernel density estimation (KDE) of the Gini coefficients of the selected training samples, or minimizing the earth-mover’s distance (EMD) between the distribution of the test dataset Gini coefficients and the Gini coefficients of the selected training samples. The random sampling, KDE distribution sampling, and EMD distribution sampling methods all sought to mimic the distribution of the test dataset for each class. The random sampling method achieved its objective via the Central Limit Theorem \cite{fullersamplingstatistics}, whereas the KDE\footnote{https://docs.scipy.org/doc/scipy/tutorial/stats/kernel\_density\_estimation.html (accessed 14Oct2024).} and EMD\footnote{https://docs.scipy.org/doc/scipy/reference/generated/scipy.stats.wasserstein\_distance.html (accessed 14Oct2024).} approaches were iterative, with 1,000 iterations per instance, to achieve the strongest alignment to the desired distributional target. 

\begin{figure}[ht]
\begin{center}
\includegraphics[width=100mm]{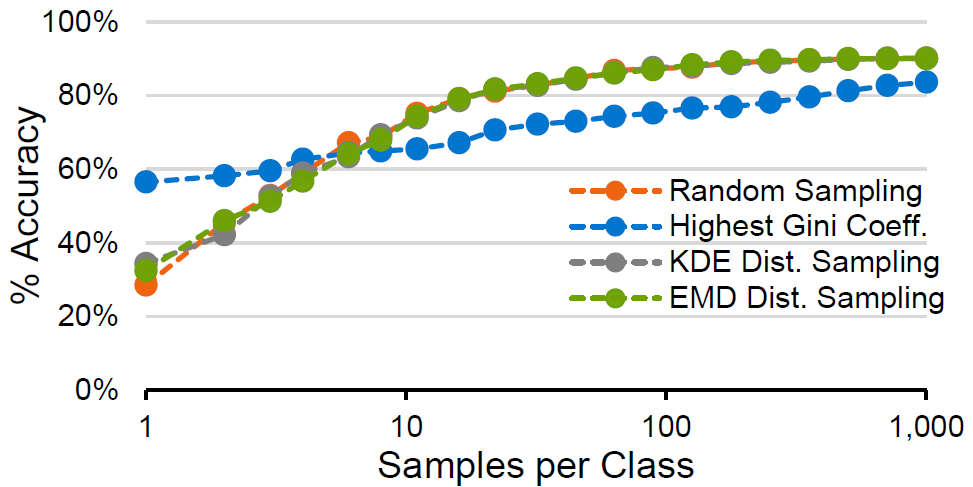}
\caption{Log-linear plot showing the accuracy of support vector machine (SVM) multiclass classification of the 10,000-example MNIST test dataset versus the number of samples per class from the 60,000-example MNIST training dataset used for training the SVM. The training dataset samples were selected using one of the following methods: per-class random sampling (orange), prioritizing samples with the highest per-class Gini coefficients (blue), mimicking the per-class distribution of the test dataset Gini coefficients using kernel density estimation (KDE) of the Gini coefficients of the selected training samples (gray), or minimizing the earth-mover’s distance (EMD) between the distribution of the test dataset Gini coefficients and the Gini coefficients of the selected training samples (green). The mean of three random sampling results is shown for each "random sampling" data point (orange). All MNIST inputs to the SVM model were flattened raw pixel values (dimension $d=784$) ranging from 0 to 255 for each $28 \times 28$ grayscale image. Each $d$-dimensional image vector was $\ell_2$-normalized before entering the SVM model.}
\end{center}
\end{figure}

The results of these different sampling strategies are shown in Figure A5, and illustrate that the random, KDE distribution, and EMD distribution sampling strategies produced essentially identical outcomes. Further, prioritizing the samples with the highest per-class Gini coefficients outperformed all other sampling strategies when selecting only one or two exemplary samples per class.

Next, we evaluated impact of random, KDE distribution, and EMD distribution sampling of the MNIST training dataset on the Gini coefficients of the selected samples associated with each method. In this analysis, Gini coefficients of the target distribution (\emph{i.e.}, test dataset) for a single class of the MNIST testing dataset were plotted against the selected samples from the MNIST training dataset for the same class. It was noteworthy that the distribution of the Gini coefficients of the sampled training data essentially mimicked the distributions of the Gini coefficients from test datasets (Figure A6). These results, taken with the SVM classifier results from Figure A5, suggested that the Gini coefficient can also be useful in assessing the diversity of a dataset distribution.

\begin{figure}[ht]
\begin{center}
\includegraphics[width=150mm]{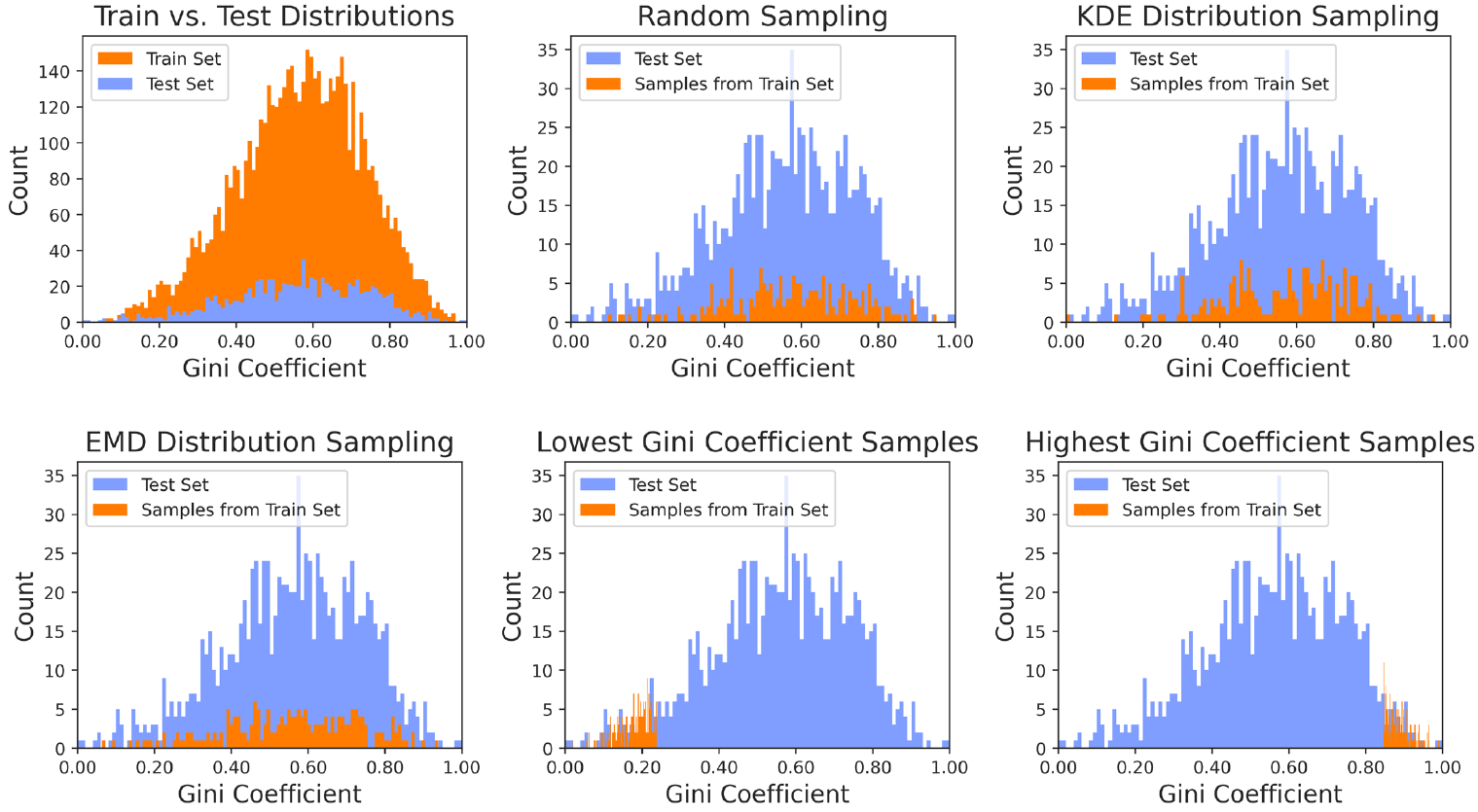}
\caption{Binned histograms (bins = 100) of Gini coefficient distributions for 200 samples drawn from the MNIST training dataset (orange) versus the MNIST test dataset (blue) for MNIST class = 2. As a baseline, the full training set versus full test set distributions for class = 2 are shown on the top-left plot. The training dataset samples were selected using one of the following methods: random sampling (top-middle), mimicking the distribution of the test dataset Gini coefficients using kernel density estimation (KDE) of the Gini coefficients of the selected training samples (top-right), minimizing the earth-mover’s distance (EMD) between the distribution of the test dataset Gini coefficients and the Gini coefficients of the selected training samples (bottom-left), prioritizing the samples with the lowest Gini coefficients (bottom-middle), or prioritizing the samples with the highest Gini coefficients (bottom-right). Note: the bin sizes were held constant (bins = 100) for all plots to maintain consistency. The lowest/highest Gini coefficient sampling plots (bottom-middle and bottom-right) have a narrow training sample distribution, which makes the bars for the sample bins appear very slim. However, the overall sample count for each plot is consistent with the other plots in the figure.}
\end{center}
\end{figure}

\newpage

\subsection{Classifier Accuracies for Fashion-MNIST Dataset Images}

We explored the role of sampling strategies when training a support vector machine (SVM) machine learning (ML) algorithm. Training datasets were sampled with increasing numbers of samples per class in the multiclass dataset. The resulting trained ML models were evaluated using a hold-out testing dataset that was consistent across the entire analysis. 

In our study, the training dataset samples were selected from the 60,000-example Fashion-MNIST training dataset by either random sampling, prioritizing the samples with the highest per-class Gini coefficients (\emph{i.e.}, most exemplary and least diverse samples), or prioritizing the samples with the lowest per-class Gini coefficients (\emph{i.e.}, most diverse samples). The mean of three random sampling results were used to generate each random sampling data point. All Fashion-MNIST training and testing data inputs to the SVM model were flattened raw pixel values ($d = 784$) ranging from 0 to 255 for each $28 \times 28$ grayscale image. Each $d$-dimensional image vector was $\ell_2$-normalized before entering the SVM model for either training or testing. SVM model performance was assessed by the classification accuracy across all classes for the test dataset for each training sample size cohort.

We noted that when selecting only one or two exemplary samples per class, prioritizing the samples with the highest per-class Gini coefficients outperformed all other sampling strategies (Figure A7). Conversely, prioritizing the samples with the lowest per-class Gini coefficients resulted in significantly lower performance than the other sampling methods. In aggregate, the random sampling approach outperformed the other methods and led to near-optimal performance of the SVM classifier with undersampled training dataset examples. All methods converged to the maximum accuracy of 84\% as they neared the maximum number of approximately 6,000 samples per class of the Fashion-MNIST training data set due to the Central Limit Theorem \cite{fullersamplingstatistics}

\begin{figure}[ht]
\begin{center}
\includegraphics[width=100mm]{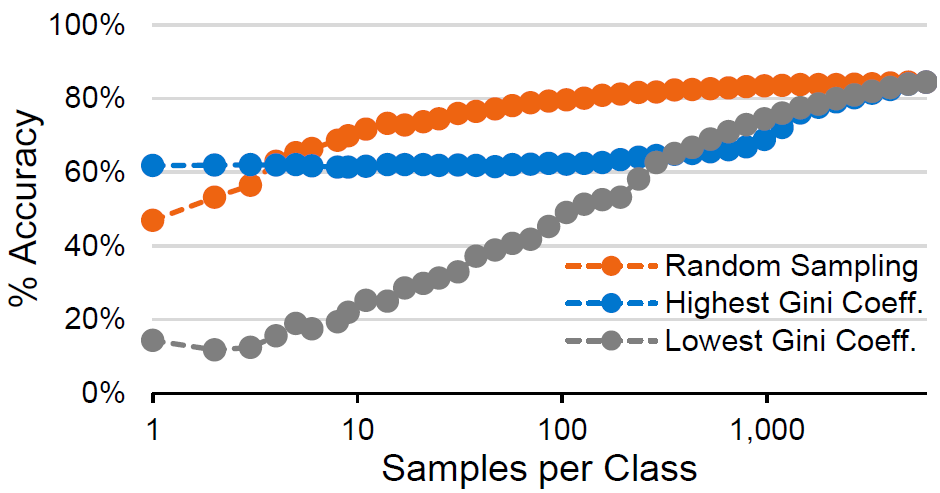}
\caption{Log-linear plot showing the accuracy of support vector machine (SVM) multiclass classification of the 10,000-example Fashion-MNIST test dataset versus the number of samples per class from the 60,000-example Fashion-MNIST training dataset used for training the SVM. The training dataset samples were selected by either per-class random sampling (orange), prioritizing the samples with the highest per-class Gini coefficients (blue), or prioritizing the samples with the lowest per-class Gini coefficients (gray). The mean of three random sampling results is shown for each "random sampling" data point (orange). All Fashion-MNIST inputs to the SVM model were flattened raw pixel values (dimension $d=784$) ranging from 0 to 255 for each $28 \times 28$ grayscale image. Each $d$-dimensional image vector was $\ell_2$-normalized before entering the SVM model.}
\end{center}
\end{figure}

As we also observed with the MNIST dataset, prioritizing the top one or two samples with the highest per-class Gini coefficients from the Fashion-MNIST dataset resulted in the selection of exemplary samples from each class. This approach outperformed random selection of training samples and was particularly valuable when only one or two samples per class are used for multiclass ML training campaigns.

At a high level, we observed that selecting machine learning training samples that closely matched the distribution of the testing dataset through random sampling was much more important than ensuring data diversity (\emph{i.e.}, samples with lower Gini coefficients). Selection of exemplary and iconic training samples (\emph{i.e.}, samples with higher Gini coefficients) led to significantly better model performance compared to simply having a diverse training set (\emph{i.e.}, samples with lower Gini coefficients). These consistent results further highlighted the effectiveness of our method in prioritizing samples with the highest per-class Gini coefficients, especially in situations with very sparse information.

\end{document}